\documentclass[11pt,a4paper]{article}
\usepackage[hyperref]{eacl2021}
\usepackage{times}
\usepackage{latexsym}
\usepackage{listings}
\usepackage{booktabs}
\usepackage{color,soul}
\usepackage{multirow}
\usepackage{todonotes}

\lstdefinestyle{Python}{
    language        = Python,
    frame           = lines, 
    basicstyle      = \footnotesize,
    stringstyle     = \color{green},
    commentstyle    = \color{red}\ttfamily
}

\usepackage{microtype}

\aclfinalcopy 

\title{Nearest neighbour approaches for Emotion Detection in Tweets}

\author{Olha Kaminska
\\Computational Web Intelligence 
  \\Department of Applied Mathematics,\\ Computer Science and Statistics \\ Ghent University
  \And
  Chris Cornelis
  \\Computational Web Intelligence 
  \\Department of Applied Mathematics,\\ Computer Science and Statistics \\ Ghent University 
  \AND Veronique Hoste \\ LT3 Language and Translation Technology Team \\ Ghent University \\\texttt{\{Olha.Kaminska,Chris.Cornelis,Veronique.Hoste\}@UGent.be}}
\date{}

\begin{document}
\maketitle
\begin{abstract}
Emotion detection is an important task that can be applied to social media data to discover new knowledge. While the use of deep learning methods for this task has been prevalent, they are black-box models, making their decisions hard to interpret for a human operator. Therefore, in this paper, we propose an approach using weighted $k$ Nearest Neighbours (kNN), a simple, easy to implement, and explainable machine learning model. These qualities can help to enhance results' reliability and guide error analysis. \par 
In particular, we apply the weighted kNN model to the shared emotion detection task in tweets from SemEval-2018. Tweets are represented using different text embedding methods and emotion lexicon vocabulary scores, and classification is done by an ensemble of weighted kNN models. Our best approaches obtain results competitive with state-of-the-art solutions and open up a promising alternative path to neural network methods. 
\end{abstract}

\section{Introduction}
\label{intro}

In this paper, we consider SemEval-2018 Task 1 EI-oc: Affect in Tweets for English\footnote{\url{https://competitions.codalab.org/competitions/17751}} \cite{SemEval2018Task1}. This is a classification problem in which data instances are raw tweets, labeled with scores expressing how much each of four considered emotions (anger, sadness, joy, and fear) are present. \par 
Our target is to implement the weighted $k$ Nearest Neighbor (wkNN) algorithm to detect emotions in tweets. In doing so, we consider different ways of tweet embeddings and combine them with various emotional lexicons, which provide an emotional score for each word. \par 
The motivation for using wkNN is to show the potential of a simple, interpretable machine learning approach compared to black-box techniques based on more complex models like neural networks (NNs). By contrast to the latter, wkNN's predictions for a test sample can be traced back easily to the training samples (the nearest neighbours) that triggered this decision.\par
We note that we still use NN-based methods for obtaining tweet embeddings. One could therefore argue that our method is not fully explainable; however, we feel that it is less important to understand how tweets are initially represented in an $n$-dimensional space, than to explain how they are used in making predictions for nearby instances. \par
The remainder of this paper is organized as follows: in Section \ref{works} we discuss related work, mainly focusing on the winning approaches of SemEval-2018 Task 1. In Section \ref{method}, we describe the methodology behind our solution, including data cleaning, tweet representations through word embeddings, lexicon vocabularies, and their combinations; our proposed ensemble method for classification; and finally, evaluation measures. In Section \ref{experiments}, we report the observed performance on training and development data for the different setups of our proposal, while Section \ref{results} lists the results of the best approach on the test data and compares them to the competition results. In Section \ref{discussion}, we examine some of the test samples with correct and wrong predictions to see how we can use our model's interpretability to explain the obtained results. Finally, in Section \ref{conclusion}, we discuss our results and consider possible ways to improve them.

\section{Related work}
\label{works}

First, we briefly recall the most successful proposals\footnote{Competition results: \url{https://competitions.codalab.org/competitions/17751\#results}} to the SemEval-2018 task. The winning approach \cite{duppada2018seernet} uses tweet embedding vectors in ensembles of XGBoost and Random Forest classification models. The runners-up \cite{gee2018psyml} perform transfer learning with Long Short Term Memory (LSTM) neural networks. The third-place contestants \cite{rozental2018} train an ensemble of a complex model consisting of Gated-Recurrent-Units (GRU) using a convolution neural network (CNN) as an attention mechanism. \par 
It is clear that the leaderboard is dominated by solutions that are neither simple nor interpretable. This comes as no surprise, given that the effectiveness of a solution is evaluated only using the Pearson Correlation Coefficient (see formula (\ref{eq:pcc}) in Section 3.5).\par 
In general, machine learning models in the Natural Language Processing (NLP) field rarely explain their predicted labels. This inspires the need for explainable models, which concentrate on interpreting outputs and the connection of inputs with outputs. For example, \citet{liu2018towards} present an explainable classification approach that solves NLP tasks with comparable accuracy to neural networks and also generates explanations for its solutions. \par
Recently, \citet{danilevsky2020survey} presented an overview of explainable methods for NLP tasks. Apart from focusing on explanations of model predictions, they also discuss the most important techniques to generate and visualize explanations. The paper also discusses evaluation techniques to measure the quality of the obtained explanations, which could be useful in future work. \par 
In this paper, we consider one of the simplest explainable models: the kNN method. In the context of NLP, kNN has recently been applied by \cite{rajani2020explaining} as a backoff method for classifiers based on BERT and RoBERTa (see Section \ref{section:TweetEmbedding}). In particular, when the latter NN methods are less confident about their predictions, the kNN solution is used instead. In this paper, we will only use such NN approaches at the data representation level and rely on weighted kNN only during classification.

\section{Methodology}
\label{method}

In this section, we describe the different ingredients of our approach, more precisely, data preprocessing, embedding methods, emotional lexicon vocabularies, classification, and evaluation methods.\par 
We focus on the emotion intensity ordinal classification task (EI-oc) \cite{SemEval2018Task1}. Given each of the four considered emotions (anger, fear, joy, sadness), the task is to classify a tweet in English into one of four ordinal classes of emotion intensity (0: no emotion can be inferred, 1: low amount of emotion can be inferred, 2: moderate amount of emotion can be inferred, 3: high amount of emotion can be inferred) which best represents the mental state of the tweeter.\par 
Separate training, development, and test datasets were provided for each emotion. To train the classification model, we merge the training and development datasets to evaluate our results with the cross-validation method. \par

\subsection{Data cleaning}

Before starting the embedding process, we can clean tweets in several ways: 
\begin{itemize}
    \item General preprocessing. First, we delete account tags (starting with $'@'$), newline symbols ('$\backslash$n'), extra white spaces, all punctuation marks, and numbers. Next, we replace '$\&$' with the word 'and' and replace emojis with textual descriptions.\par 
    We save hashtags as a potential source of useful information \cite{mohammad2015using} but delete $'\#'$ symbols.\par
We do not delete emojis because, following the observations from \citet{wolny2016emotion}, using emoji symbols could significantly improve precision in identifying various types of emotions. In the source data, emojis are present in two ways: combinations of punctuation marks and/or letters and small pictures decoded with Unicode. The first type of emojis is replaced with their descriptions taking from the list of emoticons on Wikipedia\footnote{\url{https://en.wikipedia.org/wiki/List\_of\_emoticons}}. The second type of emojis are transformed using the Python package \textit{"emoji"} \footnote{\url{https://pypi.org/project/emoji/}}.\par

\item Stop-word removal: for this process, the list of stop-words from the NLTK package\footnote{\url{http://www.nltk.org/nltk_data/}} is used. 
\end{itemize}

We do not apply preprocessing or stop-word removal a priori, but rather examine whether they improve the classification during the experimental stage. 

\subsection{Tweet embedding}
\label{section:TweetEmbedding}

To perform classification, each tweet is represented by a vector or set of vectors, using the following word embedding techniques:
\begin{itemize}
    \item Pre-trained Word2Vec from the Gensim package\footnote{\url{https://radimrehurek.com/gensim/models/word2vec.html}}. This model includes 300-dimension word vectors for a vocabulary with 3 million words and phrases trained on a Google News dataset. It is included here because of its popularity in NLP tasks.
    \item DeepMoji\footnote{\url{https://deepmoji.mit.edu/}} is a state-of-the-art sentiment embedding model, pre-trained on millions of tweets with emojis to recognize emotions and sarcasm. We used its implementation on PyTorch by Huggingface\footnote{\url{https://github.com/huggingface/torchMoji}}, which provides for each sentence an embedding of size 2304 dimensions. 
    \item The Universal Sentence Encoder (USE) \cite{cer2018universal} is a sentence-level embedding approach developed by the TensorFlow team\footnote{\url{https://www.tensorflow.org/hub/tutorials/semantic\_similarity\_with\_tf\_hub\_universal\_encoder}}. It provides a 512-dimensional vector for a sentence or even a whole paragraph that can be used for different tasks such as text classification, sentence similarity, etc. USE was trained with a deep averaging network (DAN) encoder on several data sources. \par 
    The model is available in two options: trained with a DAN and with a Transformer encoder. After basic experiments, we chose the second one for further experiments.  
    \item Bidirectional Encoder Representations from Transformers (BERT) by \citet{devlin2018bert}. The used script\footnote{\url{https://github.com/dnanhkhoa/pytorch-pretrained-BERT/blob/master/examples/extract_features.py}} was developed by The Google AI Language Team and extracted pre-computed feature vectors from a PyTorch BERT model. The length of the output vector for a word is 768 features. Words that are not in the BERT vocabulary were split into tokens (for example, the word "$tokens$" will be resented as "$tok$", "$\#\#en$", "$\#\#s$"), and for each token, a vector was created. 
    \item Sentence-BERT (SBERT) is a modified and tuned BERT model presented in \citet{reimers2019sentence}. It uses so-called siamese and triplet network structures, or a ``twin network", that processes two sentences in the same way simultaneously. SBERT provides embeddings at a sentence level with the same size as the original BERT. 
    \item Twitter-roBERTa-based model for Emotion Recognition, one of the seven fine-tuned roBERTa models presented by \citet{barbieri2020tweeteval}. Each described model was trained for a specific task and provided an embedding at the token level similar to BERT. The model that we consider was trained for the emotion detection task (E-c) using a different collection of tweets from the same authors of SemEval 2018 Task 1 \cite{SemEval2018Task1}, in which the emotions anger, joy, sadness, and optimism are used.
    
\end{itemize}
Sentence-level embeddings are applied to each tweet as a whole, while for word (or token) level embeddings, we represent a tweet vector as the mean of its words' (tokens') vectors. 

\subsection{Emotional lexicon vocabularies}

As an additional source of information to complement tweet embeddings, we also consider lexicon scores. Emotional lexicons are vocabularies that provide scores of different emotion intensity for a word. In our experiments, we use the following English lexicons:
\begin{itemize}
    \item Valence Arousal Dominance (NRC VAD) lexicon (20,007 words) \cite{vad-acl2018} – each word has a score (float number between 0 and 1) for Valence, Arousal, and Dominance.
    \item Emotional Lexicon (EMOLEX) (14,182 words) lexicon \cite{Mohammad13} – each word has ten scores (0 or 1), one per emotion: anger, anticipation, disgust, fear, joy, negative, positive, sadness, surprise, and trust.
    \item Affect Intensity (AI) lexicon (nearly 6,000 terms) \cite{LREC18-AIL} – each word has four scores (float number from 0 to 1), one per emotion: anger, fear, sadness, and joy.
    \item Affective norms for English words (ANEW) lexicon (1034 words) \cite{bradley1999affective} – each word has six scores (float number between 0 and 10): Mean and SD for Valence, Arousal and Dominance.
    \item Warriner’s lexicon (13,915 lemmas) \cite{warriner2013norms} – each word has 63 scores (float number between 0 and 1000 ), reflecting different statistical characteristics of Valence, Arousal, and Dominance.
\end{itemize}

We consider the following two methods of combining word embeddings with lexicon vocabulary scores:
\begin{itemize}
    \item For each word during the embedding process, lexicon scores are appended to the end of the tweet vector. The size of the obtained vector is the word embedding size plus the number of lexicon scores.
    \item We construct a separate feature for each lexicon. These models are then combined with the embedding vectors in an ensemble classifier, as described in Section 3.4.
\end{itemize}
We perform experiments for all emotion datasets with one or several lexicons. The results are presented in Section 4.

\subsection{Classification methods} 

In this subsection, the weighted $k$ Nearest Neighbors (wkNN) classification method \cite{dudani1976distance} and its similarity relation are described. \par 
The wkNN is a refinement of the regular kNN, where distances to the neighbors are taken into account as weights. This approach aims to assign more significant weight to the closest instances and a smaller weight to the ones that are further away. The wkNN has two main parameters: the used metric or similarity relation and the number $k$ of considered neighbours.\par
To choose an appropriate similarity relation, we follow \citet{huang2008similarity}, who compared metrics for the document clustering task. The cosine metric was shown to be one of the best:
\begin{equation}
\label{eq:cos}
cos(A, B) = \frac{A \cdot B}{||A|| \times ||B||},
\end{equation}
where $A$ and $B$ denote elements from the same vector space, $A \cdot B$ is their scalar product, $||*||$ - vector norm. \par 
Values provided by this measure are between -1 (perfectly dissimilar vectors) and 1 (perfectly similar vectors). In order to obtain a [0,1]-similarity relation instead of a metric, we use the following formula:
\begin{equation}
\label{eq:cos_sim}
cos\_similarity(A, B) = \frac{1 + cos(A, B)}{2}.
\end{equation}

Formula (\ref{eq:cos_sim}) is used as the primary similarity relation throughout this paper.

Regarding the parameter $k$, there is no one-fits-all rule to determine it. As a general \textit{"rule of thumb"}, we can put $k =\frac{\sqrt{N}}{2}$, where $N$ is the number of samples in the dataset. However, to examine the impact of $k$, we will use various numbers of neighbors for each emotion dataset for the best-performing methods in our experiments. \par 

We use wKNN both as a standalone method as well as inside of a classification ensemble. For the latter, a separate model is trained for each information source (vectors containing tweet embeddings, lexicon scores, or their combination).\par
\begin{figure}[!ht]
\centering
  \includegraphics[width=1\linewidth]{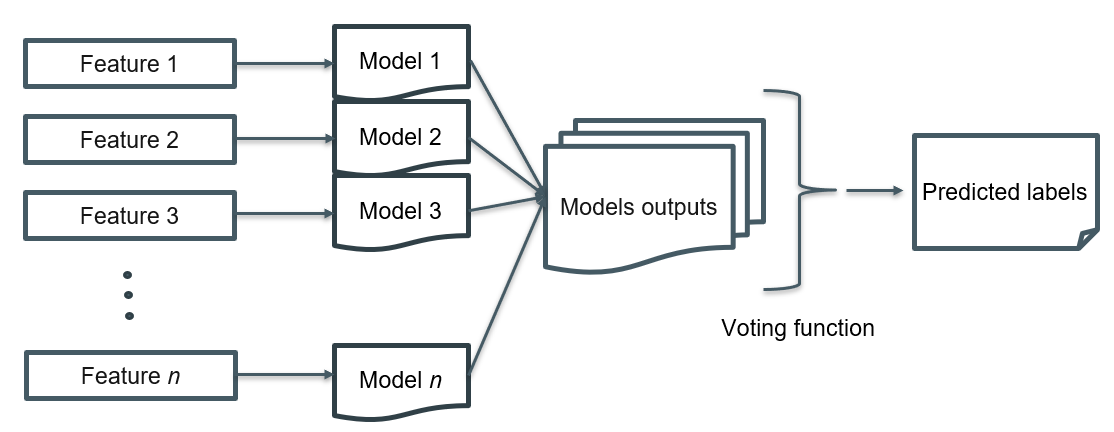}
  \caption{The scheme of the ensemble architecture.}
  \label{fig:ensemble}
\end{figure}

 
For each test sample, each model's outputs are combined using the standard average as a voting function, i.e., each model gets the same weight in this vote. The architecture of our approach is illustrated in Fig. \ref{fig:ensemble}. \par 
Note that in this way, the predictions will be float values between 0 and 3, rather than integer labels (0, 1, 2 or 3); however, at the training stage, this does not represent a problem.

\begin{table*}[!ht]
\centering
\caption{The best setup for each emotion for different embeddings.}
\label{tab:embed_wknn_best}
\renewcommand{\arraystretch}{0.8}
\begin{tabular}{l|cccc}
    \toprule[1pt]\midrule[0.3pt]
    \bf{Setup} & \bf{Anger} & \bf{Joy} & \bf{Sadness} & \bf{Fear} \\
    \midrule\midrule
    \multicolumn{5}{c}{\bf{roBERTa-based}}\\
    Tweets preprocessing & No & No & No & Yes\\
    Stop-words cleaning & No & No & No & No \\
    Number of neighbors & 19 & 13 & 9 & 11\\ 
    PCC & \bf{0.6651} & \bf{0.6919} & \bf{0.7055} & 0.5694\\
    \midrule
    \multicolumn{5}{c}{\bf{DeepMoji}}\\
    Tweets preprocessing & Yes & Yes & Yes & Yes\\
    Stop-words cleaning & Yes & No & No & No\\
    Number of neighbors & 11 & 21 & 13 & 13\\ 
    PCC & 0.6190 & 0.6426 & 0.6490 & \bf{0.5737}\\
    \midrule
    \multicolumn{5}{c}{\bf{USE}}\\
    Tweets preprocessing & Yes & Yes & Yes & No\\
    Stop-words cleaning & No & No & No & No \\
    Number of neighbors & 19 & 21 & 19 & 11\\ 
    PCC & 0.5174 & 0.5580 & 0.6067 & 0.5589\\
    \midrule
    \multicolumn{5}{c}{\bf{SBERT}}\\
    Tweets preprocessing & Yes & Yes & Yes & Yes\\
    Stop-words cleaning & No & No & No & No \\
    Number of neighbors & 21 & 9 & 21 & 13\\ 
    PCC & 0.4946 & 0.5413 & 0.5505 & 0.4608\\
    \midrule
    \multicolumn{5}{c}{\bf{Word2Vec}}\\
    Tweets preprocessing & Yes & Yes & Yes & Yes\\
    Stop-words cleaning & Yes & Yes & Yes & Yes \\
    The number of neighbors & 5 & 23 & 21 & 13\\ 
    PCC & 0.4824 & 0.4791 & 0.5136 & 0.4303\\
    \midrule[0.3pt]\bottomrule[1pt]
\end{tabular}
\end{table*}

\subsection{Evaluation method}

To evaluate the performance of the implemented methods, $5$-fold cross-validation is used, using as evaluation measure the Pearson Correlation Coefficient (PCC), as was also done for the competition. \par 
Given the vectors of predicted values $y$ and correct values $x$, the PCC measure provides a value between $-1$ (a total negative linear correlation) and $1$ (a total positive linear correlation), where $0$ represents no linear correlation.\par
Hence, the best model should provide the highest value of PCC:
\begin{equation}
\label{eq:pcc}
PCC = \frac{\sum_i{(x_i-\bar{x})(y_i-\bar{y})}}{\sqrt{\sum_i{(x_i-\bar{x})^2}\sum_i{(y_i-\bar{y})^2}}}.
\end{equation}
Here $x_i$ and $y_i$ refer to the ${i}^{th}$ component of vectors $x$ and $y$, while $\bar{x}$ and $\bar{y}$ represent their mean.\par
The correlation scores across all four emotions were averaged by the competition organizers to determine the bottom-line metric by which the submissions were ranked. \par 

\section{Experiments}
\label{experiments}

Our experiments on the train and development are designed as follows: first, we compare the individual tweet embedding methods (Section 4.1) and examine which setup gives the best results. Then, in Section 4.2, we also involve the emotional lexicons, either independently, by appending them to tweet embedding vectors and in ensembles.

\subsection{Detecting the best setup for embeddings}

In this subsection, we describe the process of detecting the best data cleaning method and the best $k$ parameter value for each emotion dataset and each embedding. The results are shown in Table \ref{tab:embed_wknn_best}. \par 
In the first step, for each emotion and each embedding, we calculate the PCC for different versions of the dataset: original raw tweets, preprocessed tweets, and preprocessed tweets with stop-words removed. To verify which approach works better, we perform statistical analysis using the two-sided $t$-test in Python's package $'stats'$.\par 
In the second step, we repeat the experiments for the best preprocessing setups with different amounts of neighbours $(5, 7, 9, ..., 23)$ to detect the most appropriate $k$ value. These values and the resulting PCC for the optimal setup are shown in Table \ref{tab:embed_wknn_best}.\par
We can observe that stop-word cleaning only improved results for the Word2Vec embedding and that for the roBERTa-based model, it makes sense to use the raw tweets. \par 
Also, among the different embeddings, roBERTa obtains the highest results on three out of four datasets, while for Fear, the best result is obtained by DeepMoji (with roBERTa a close second). This can be explained by the fact that these two embeddings are explicitly trained on emotion data. The three remaining embeddings lag considerably, with the notable exception of USE for Fear. We conjecture that this may have to do with the imbalanced nature of the fear dataset. \par 
To measure the balance of the different datasets, we calculated the Imbalance Ratio (IR) of the combined train and development data, where IR is equal to the ratio of the sizes of the largest and smallest classes in the dataset. A value close to 1 represents balanced data. \par 
With IR values of $1.677$ and $1.47$, respectively, the anger and joy datasets can be considered fairly balanced. While the imbalance is somewhat higher for the sadness dataset (IR = $2.2$), the fear dataset is the most imbalanced dataset among with a IR value of $8.04$.

\subsection{Combining embeddings and lexicons}

In this subsection, we discuss our experiments joining the previously identified best setups of the embedding methods with all emotional lexicons, using the different combination strategies outlined in Section 3.3. \par

\subsubsection{Lexicon-only based models}

We first evaluate models based purely on lexicons. The goal here is to check the intrinsic classification strength of each lexicon and of the lexicon-based approach as a whole. \par 
A lexicon works as a dictionary: if a word is present in the lexicon, it receives a particular score, in the other case, it is assigned a score of zero. For lexicons with several scores per word, we take all of them. To obtain the lexicon score for a full tweet, as usual, we compute the mean of its words' scores. \par 
For each of the five lexicons, the output is saved as a separate vector. The sixth vector is constructed by combining all lexicons' scores and has a total length of 86 values (the sum of the number of scores for all five lexicons). For each of these vectors, a weighted kNN classification model is applied. \par 
Initially, we use the same number of neighbours for all datasets, computed using the rule of thumb $k$ = $\sqrt{N}/2$, where $N$ is the size of the dataset. The dataset sizes are mostly near to 2000 instances, so $k$ = 23 is used. Results are presented in Table \ref{tab:lex}.

\begin{table}[!ht]
\centering
\caption{Results for the lexicon-based approach.}
\label{tab:lex}
\renewcommand{\arraystretch}{0.7}
\begin{tabular}{l|cccc}
	\toprule[1pt]\midrule[0.3pt]
	\bf{Lexicon} & \bf{Anger} & \bf{Joy} & \bf{Sadness} & \bf{Fear}\\
	\midrule\midrule
	VAD&0.1983& 0.2823&0.2043&0.0928\\
	EMOLEX&0.3014&0.2893&0.3404&0.1943\\
	AI&\bf{0.3284}&0.2673&\bf{0.3723}&0.1549\\
	ANEW&0.1972&0.3050&0.3254&\bf{0.2278} \\
	Warriner& 0.1901&0.2705&0.2970&0.1505\\
	Combined&0.2133&\bf{0.3051}&0.3151&0.1626\\
	\midrule[0.3pt]\bottomrule[1pt]
\end{tabular}
\end{table}
We can observe that the AI lexicon is the best performing lexicon, showing the highest results for two out of the four datasets. \par
Then, for each emotion dataset and its best performing lexicon, the best $k$ value is detected. These results are presented in Table \ref{tab:lex_setup}. 

\begin{table}[!ht]
\centering
\caption{The best setup for each emotion for different lexicon-based feature vectors.}
\label{tab:lex_setup}
\renewcommand{\arraystretch}{0.8}
\begin{tabular}{l|cccc}
    \toprule[1pt]\midrule[0.3pt]
    \bf{Dataset} & \bf{Lexicon} & \bf{$k$ value} & \bf{PCC} \\
    \midrule\midrule
    Anger & AI & 11 & 0.3359\\
    Joy & Combined & 19 & 0.3320\\
    Sadness & AI & 23 & 0.3723\\ 
    Fear & ANEW & 17 & 0.2412\\
    \midrule[0.3pt]\bottomrule[1pt]
\end{tabular}
\end{table}

As we can see, for different datasets, different values of $k$ perform better. \par 

\begin{table}[!ht]
\centering
\caption{Results for the first combination approach.}
\label{tab:embedd_lex1}
\renewcommand{\arraystretch}{0.7}
\begin{tabular}{l|cccc}
	\toprule[1pt]\midrule[0.3pt]
	\bf{Lexicon} & \bf{Anger} & \bf{Joy} & \bf{Sadness} & \bf{Fear}\\
	\midrule\midrule
	\multicolumn{5}{c}{\bf{roBERTa-based model}}\\
	VAD& 0.6536&0.6845&0.6986&0.5424\\
	EMOLEX&\bf{0.6653}&0.6819&\bf{0.7106}&\bf{0.5712}\\
	AI& 0.6636&0.6853&0.6996&0.5582\\
	ANEW& 0.6573&\bf{0.6927}&0.6859&0.5505\\
	Warriner& 0.6519&0.6757&0.6969&0.5563\\
	None& 0.6651&0.6919&0.7055&0.5694\\
	\midrule\midrule
	\multicolumn{5}{c}{\bf{DeepMoji}}\\
	VAD&0.6060&0.6292&0.6380&0.5557\\
	EMOLEX&0.5923&0.6235&0.6449&0.5672 \\
	AI& 0.6124&0.6269&0.6325&0.5754\\
	ANEW&0.6046&0.6352&0.6356&0.5613\\
	Warriner& 0.6002&0.6215&0.6325&\bf{0.5795}\\
	None&\bf{0.6190}&\bf{0.6426}&\bf{0.6490}&0.5737\\
	\midrule\midrule
	\multicolumn{5}{c}{\bf{USE}}\\
	VAD&0.5079&0.5526&0.5881&0.5382\\
	EMOLEX&0.5042&0.5509&0.5786&0.5451 \\
	AI& 0.4932&0.5462&0.5961&0.5447\\
	ANEW&\bf{0.5293}&0.5484&0.5932&0.5297\\
	Warriner& 0.5071&0.5475&0.5829&0.5589\\
	None& 0.5174&\bf{0.5580}&\bf{0.6067}&\bf{0.5589}\\
	\midrule\midrule
	\multicolumn{5}{c}{\bf{SBERT}}\\
	VAD& 0.4781&0.5247&0.5249&0.4506\\
	EMOLEX& 0.4808&0.5363&0.5303&0.4512\\
	AI& 0.4631&0.5439&0.5395&0.4273\\
	ANEW& 0.4835&0.5461&\bf{0.5576}&0.4486\\
	Warriner& 0.4773&\bf{0.5481}&0.5169&0.4591\\
	None&\bf{0.4946}&0.5413&0.5505&\bf{0.4608}\\
	\midrule\midrule
	\multicolumn{5}{c}{\bf{Word2Vec}}\\
	VAD& 0.4463&0.4662&0.4887&\bf{0.4463}\\
	EMOLEX&0.4574&0.4749&0.5098&0.4215 \\
	AI& 0.4527&0.4728&0.5108&0.4247\\
	ANEW&0.4621&0.4779&\bf{0.5156}&0.4382\\
	Warriner& 0.4574&0.4716&0.5039&0.4340\\
	None&\bf{0.4824}&\bf{0.4791}&0.5136&0.4303\\
	\midrule[0.3pt]\bottomrule[1pt]
\end{tabular}
\end{table}

\begin{table*}[!ht]
\centering
\caption{Results for the ensemble approach with different feature vectors, for all datasets.}
\label{tab:features}
\renewcommand{\arraystretch}{0.8}
\begin{tabular}{lc|cccc}
    \toprule[1pt]\midrule[0.3pt]
\bf{The vectors} & \bf{Vector size} & \bf{Anger} & \bf{Joy} & \bf{Sadness} & \bf{Fear}\\
    \midrule\midrule
    The baseline (top-five embeddings vectors) & 5 & 0.6929 & 0.7420 & 0.7329 & 0.6783\\ 
    With the best lexicon & 6 & 0.6902 & 0.7336&0.7400&0.6773\\
    \shortstack{With all five lexicons\\and their combination} & 11 & 0.6431& 0.6796&0.6962&0.6585\\
    \midrule
    \shortstack{With roBERTa combined\\with the best lexicon} & 6 & 0.7120 & 0.7496 & \bf{0.7579} &  0.6719\\
    \shortstack{With the best lexicon and \\ roBERTa combined with \\the best lexicon}& 7 & \bf{0.7190} & \bf{0.7526}
    & 0.7566 & \bf{0.6804}\\
    \midrule[0.3pt]\bottomrule[1pt]
\end{tabular}
\end{table*}

\subsubsection{Models appending lexicon scores to word embeddings}

In this approach, embedding and lexicon scores are normalized to values between 0 and 1 to account for differences in ranges. To obtain the vector of a tweet, we take the average of all vectors of its words. \par
The results of these combination experiments are provided in Table \ref{tab:embedd_lex1}. To check the appending strategy's added value, Table \ref{tab:embedd_lex1} also presents the previously obtained PCC score using none of the lexicons for each embedding method. \par
As can be seen, for half of the experiments, the use of lexicons does not improve the PCC value. The roBERTa-based model is the only model that seems to benefit from the added lexicon information for each emotion dataset, although the improvement is marginal.\par
For three out of four datasets, for the roBERTa-based model, the EMOLEX lexicon was the best. For other embedding models, mostly approach with no lexicon benefited, and when some lexicons improved results, they were different for different datasets, with no noticeable pattern. If we compare the best lexicons from Table~\ref{tab:embedd_lex1} with the best ones from Table~\ref{tab:lex_setup}, we can see that they are different for each dataset.\par

\subsubsection{Ensembles}

The first ensemble that we tried combines the five classifiers based on embedding models from Section 4.1, i.e., the roBERTa-based, DeepMoji, USE, SBERT, and Word2Vec embeddings. We train the weighted kNN models for each vector separately with the best $k$ value and tweet preprocessing pipeline. Results are listed in the first line of Table \ref{tab:features} and indicate that these five embeddings already provide a good baseline, improving the best results from Table \ref{tab:embed_wknn_best} by 8\% on average. Especially for Fear, the improvement is notable (18\% up).\par 
Next, we consider the inclusion of lexicons into the ensembles. In a first setup, for each dataset, we take the best performing lexicon from Table~\ref{tab:lex_setup} and add this is as a separate classifier to the baseline ensemble. For comparison, we also consider a setup where all five lexicons and their combination are added as six more classifiers, to check how each of them influences the output scores. The obtained results, shown in the second and third lines of Table \ref{tab:features}, illustrate that, in general, the lexicons are unable to improve the baseline and that adding all lexicons takes the scores down considerably. \par 
Given that the roBERTa-based method performs the best among all embeddings (Table \ref{tab:embed_wknn_best}), and that it is the only one that benefits from the lexicon appending strategy (Table \ref{tab:embedd_lex1}), we also consider two additional setups. One that extends the baseline with the lexicon-appended roBERTa classifier and another one that adds the best lexicon to the previous ensemble. \par
The last two approaches results are presented in the second half of Table \ref{tab:features}. We can see that these adjustments improve the scores noticeably. For three out of four emotion datasets, the last setup performs best, while for Sadness, the results are almost equal. Therefore, we consider this last ensemble as the best solution.

\section{Results on the test data}
\label{results}

To determine the generalization strength of the obtained best approach from Section 4, we evaluated it on the test data by submitting predicted labels in the required format to the competition page\footnote{\url{https://competitions.codalab.org/competitions/17751\#learn_the_details-evaluation}}. PCC scores were calculated for each emotion dataset, and obtained results were averaged. Because of the mean voting function in our model's ensemble, our predicted labels are in float format. Therefore, to match the requested format on the competition page, before submitting results, we rounded them to the nearest integer label. \par 
The obtained PCC scores are shown in Table~\ref{tab:test}, together with the results obtained on the training and development data for comparison. \par 

\begin{table}[!ht]
\centering
\caption{Pearson Coefficient of the best approach on the cross-validation and test data for the four emotion datasets.}
\label{tab:test}
\renewcommand{\arraystretch}{0.8}
\begin{tabular}{l|cc}
    \toprule[1pt]\midrule[0.3pt] 
\bf{Dataset} & \bf{Training and} & \bf{Test data}
\\
& \bf{development data} &\\
    \midrule\midrule
    Anger& 0.719&0.638\\
    Joy&0.752&0.631 \\
    Sadness&0.756&0.670\\
    Fear &0.680&0.601\\
    \midrule
    Averaged scores&0.726&0.635 \\
    \midrule[0.3pt]\bottomrule[1pt]
\end{tabular}
\end{table}

As expected, the average PCC for the test data drops several points compared to the training and development data, but, in general, our proposal appears to generalize well to new data.\par 
We also mention that the first three contestants of the SemEval 2018 competition obtained a PCC equal to 0.695, 0.653 and 0.646, respectively\footnote{\url{https://competitions.codalab.org/competitions/17751\#results}}, and that our proposal would therefore be just behind them in fourth position. 

\section{Discussion and error analysis}
\label{discussion}

To illustrate our approach's explainability, in this section we explore some correctly and wrongly predicted test samples. \par   
As an example of a correct prediction, we can take a look at an anger test tweet: {\em ``I know you mean well, but I'm offended. Prick."} with real anger class $``2"$. Our best model predicted label $2.4$, which was rounded to $2$, so our result is correct. To analyze how this label is obtained, we look at the predictions by all models separately. They are shown in Table~\ref{tab:rightsample} (sample (a)) with the number of neighbors of each class from the training data selected by each model. We can see that the roBERTa-based model was the most accurate, while most others were also close enough. \par 
Next, we also examined the neighbours chosen by the models and their classes, especially those which are selected by different models. To find some patterns, we took the intersection of the neighbours closest to the test instance, chosen by the ensemble's models. \par 
We should mention that those models are based on different embeddings, which may locate tweets in $n$-dimensional space differently. However, one tweet with class $``2"$ from the train data was chosen by $4$ models out of $7$ and five more tweets (four of them with class $``2"$ and one with class $``1"$) by $3$ models.\par 
A closer examination of those tweets revealed that all of them contain the word ``offended". From this, we could conclude that this word has a high emotional intensity that influences the sentence's tone.\par

\begin{table}[!ht]
\centering
\caption{Predictions of models from the ensemble for some test tweets.}
\label{tab:rightsample}
\renewcommand{\arraystretch}{0.8}
\begin{tabular}{l|c|cccc|cccc}
    \toprule[1pt]\midrule[0.3pt] 
\bf{\shortstack{Model in\\ensemble}} & \bf{k}&\multicolumn{8}{c}{\bf{Classes}}\\
&& \bf{0} & \bf{1} & \bf{2} & \bf{3}& \bf{0} & \bf{1} & \bf{2} & \bf{3} \\
\cmidrule{3-10}
 &&\multicolumn{4}{c|}{Sample (a)}& \multicolumn{4}{c}{Sample (b)}\\
    \midrule\midrule
    roBERTa&19& 0&4&11&4& 5&3&11&0\\
    DeepMoji&11& 0&0&5&6 & 2&2&5&2\\
    USE&19& 2&5&7&5& 5&2&7&5\\
    SBERT&21& 6&5&6&4& 8&8&3&2\\
    Word2Vec&5& 1&1&0&3& 0&0&3&2\\
    AI lexicon&11& 2&1&3&5& 2&5&4&0\\
    \shortstack{roBERTa\\with AI}&11& 0&2&8&1& 0&4&7&0\\
    \midrule[0.3pt]\bottomrule[1pt]
\end{tabular}
\end{table}
The next sample we examined is another anger test tweet, with gold label $``0"$: {\em ``We've been broken up a while, both moved on, she's got a kid, I don't hold any animosity towards her anymore..."} Our solution predicted a score $1.5$, which was rounded to $2$, leading to a false prediction. \par 
Similar to the previous sample, we took a look at the classes predicted by the different models in the ensemble (Table~\ref{tab:rightsample}, sample (b)). Here, we can observe that only the SBERT-based model predicted the result correctly, so roBERTa does not always provide the best answer.\par 
We also explored the most frequent neighbours, which were chosen by $3$ models (one tweet with class $``1"$) and by $2$ models (nine tweets with different classes). We did not find any noticeable patterns; the misclassification is probably caused by words with high emotional intensity, like ``animosity", which is used in combination with a negation in this specific context.

\section{Conclusion and future work}
\label{conclusion}

In this paper, we evaluate an explainable machine learning method application for the emotion detection task. As the main conclusion, we can say that using simple optimizations and the weighted kNN method can perform nearly on par with more complex state-of-the-art neural network-based approaches. In the future, we plan to incorporate more elaborate nearest neighbour methodologies, which also take into account the inherently fuzzy nature of emotion data. Some initial experiments with ordered weighted average based fuzzy rough sets \cite{cornelis2010} show promising results.\par 
Another observation that can be made from our results is that the most informative input to solving the emotion detection task is provided by the tweet embeddings, and that lexicons generally do not improve the results a lot. Meanwhile, adding the combined vector of roBERTa embedding and the best lexicon scores increased PCC scores noticeably. As a possible further improvement, we may refine the voting function by assigning different weights to the different members of the ensemble, which can be based, for example, on the confidence scores.\par 
Furthermore, as another strategy to improve results, additional text preprocessing steps could be performed, for example, using exclamation marks or word lemmatization. Also, we can give more weight to the hashtag and emoji descriptions during the tweet embedding process. \par
Another important characteristic that influences the results is data imbalance. As observed, we obtained the lowest PCC scores on the Fear dataset, most likely because it is the most imbalanced one. For further experiments with Fear, we consider the usage of imbalanced machine learning classification methods. In particular, \citet{vluymans2019dealing} discusses several approaches based on fuzzy rough set theory. \par
Finally, \citet{danilevsky2020survey} provide several hints to investigate and improve solution explainability. For example, we can examine feature importance, measure the quality of explainability, etc.

\section*{Acknowledgment}
This work was supported by the Odysseus programme of the Research Foundation—Flanders (FWO).

\bibliography{eacl2021}
\bibliographystyle{acl_natbib}

\end{document}